# VECT-GAN: A variationally encoded generative model for overcoming data scarcity in pharmaceutical science

Youssef Abdalla[1], Marrisa Taub[1], Eleanor Hilton[1], Priya Akkaraju[1], Alexander Milanovic[2], Mine Orlu[1], Abdul W. Basit[1], Michael T Cook[1], Tapabrata Chakraborti[3,4]*, David Shorthouse[1]*

[1]UCL School of Pharmacy, University College London, 29-39 Brunswick Square, London WC1N 1AX, UK
[2]Independent Researcher
[3]UCL Department of Medical Physics and Biomedical Engineering, University College London, Malet Place Engineering Building, 2 Malet Place, London WC1E 7JE, UK
[4]UCL Cancer Institute, University College London, Paul O'Gorman Building, 72 Huntley Street, London WC1E 6DD, UK

*Correspondence: t.chakraborty@ucl.ac.uk, d.shorthouse@ucl.ac.uk.

## Abstract

Data scarcity in pharmaceutical research has led to reliance on labour-intensive trial-and-error approaches for development rather than data-driven methods. While Machine Learning offers a solution, existing datasets are often small and/or noisy, limiting their utility. To address this, we developed a *Variationally Encoded Conditional Tabular Generative Adversarial Network* (VECT-GAN), a novel generative model specifically designed for augmenting small, noisy datasets. We introduce a pipeline where data is augmented before regression model development and demonstrate that this consistently and significantly improves performance over other state-of-the-art tabular generative models. We apply this pipeline across six pharmaceutical datasets, and highlight its real-world applicability by developing novel polymers with medically desirable mucoadhesive properties, which we made and experimentally characterised. Additionally, we pre-train the model on the ChEMBL database of drug-like molecules, leveraging knowledge distillation to enhance its generalisability, making it readily available for use on pharmaceutical datasets containing small molecules – an extremely common pharmaceutical task. We demonstrate the power of synthetic data for regularising small tabular datasets, highlighting its potential to become standard practice in pharmaceutical model development, and make our method, including VECT-GAN pre-trained on ChEMBL available as a pip package at: https://pypi.org/project/vect-gan/

# Introduction

Data scarcity is a major challenge for pharmaceutical research. The development of novel medicines and drug delivery methods is significantly hindered by a lack of robust datasets on which predictive models can be trained. Drugs and the vectors that deliver them to patients such as tablets, injectables, sprays, and implants are extremely costly and laborious to generate, and published examples of specific modalities often number only in the tens to hundreds[1]. The development of pharmaceuticals such as these is still typically performed in a trial and error, one-factor-at-a-time (OFAAT), or Design of Experiment (DOE) fashion[2,3], which is wasteful, time-consuming, and often unsuccessful[4]. Whilst the field has started to embrace the use of Machine Learning (ML) and Artificial Intelligence (AI) tools, these methods are highly limited by the quality and quantity of the available data[5]. Models trained and published in pharmaceutical science are often trained on only a few hundred data points, meaning that the use of powerful state-of-the-art learning methods is limited[6,7]. Furthermore, ML applied to data containing drug structures is highly sensitive to the number of available samples due to the complexity of their chemistry, with models trained on small molecule chemistry often having extremely poor accuracies compared to equivalent data in other fields. Pharmaceutical datasets also tend to display high variability due to challenges in manufacturing and characterisation, further compounding the difficulty of applying ML methods[1,8].

This challenge could be addressed by generating more data; however, this is often impractical, as a single experiment that takes weeks and costs thousands of pounds may produce only one data point[3,9]. Furthermore, these systems are often characterised by a high number of variables that can be manipulated meaning that the possible state space is vast. To address these challenges, various modelling approaches, such as DOE and OFAAT, have been explored to guide experimental design and enhance the quality of generated data[4], however, these methods still rely on the manual generation of additional data, which is not always feasible or efficient. Moreover, DOE and OFAAT are limited to modelling simple linear or quadratic relationships, which are often insufficient in the pharmaceutical field, where more complex interactions are common[2,4]. Active ML, while useful, still requires a substantial amount of data to build a final robust model[10-12]. This could be tackled using synthetic data generated from existing experimental data to help develop more regularised and robust models.

Synthetic data has been successfully applied in fields such as computer vision[13] and large language model (LLM) development[14], where it has shown great success in improving model performance by reducing the potential of a model overfitting. For example, in computer vision tasks images are augmented through techniques such as geometric transformations, kernel filtering and colour changes to enhance model robustness and symmetry[13]. Similarly, in LLM training, text perturbations, style changes and back-translation are used to reduce overfitting and increase adaptability to different domains[15]. As such, data augmentation has become standard practice in these fields. Generating synthetic data for tabular data, commonly used in pharmaceutical manufacturing, can be significantly more difficult. Tabular data often involves complex interactions and relationships between features, which cannot be easily replicated through simple perturbations. Instead, synthetic data for such applications can be created using generative models, which are trained to capture patterns in the real dataset[16]. Synthetic data has been shown to enhance model accuracy by addressing class imbalances, which are common in healthcare. For instance, Synthetic Minority Oversampling Technique (SMOTE) has been successfully used to train a Convolutional Neural Network (CNN) to detect ovarian cancer[17]. Similarly, data augmentation for underrepresented populations has proven effective in increasing the contribution of minority classes, successfully improving the detection accuracy of a rare subtype of renal cell carcinoma[18] for example. However, research on the use of synthetic data to augment small tabular datasets for subsequent ML model development remains limited.

We introduce a novel Variational Encoded Conditional Tabular–Generative Adversarial Network (VECT-GAN), specifically designed for generating synthetic tabular data common in pharmaceutical science. We present a training pipeline that leverages synthetic data generation prior to regression model training. We demonstrate how this approach significantly improves model performance across diverse use cases, applied broadly across pharmaceutical science, with VECT-GAN consistently outperforming existing state-of-the-art tabular generative models. The real-world experimental applicability of this pipeline is shown through its application and experimental validation on a small dataset of mucoadhesive polymers, which show clinical promise but are notoriously hard to manufacture and predict. Using this dataset of less than 100 samples, we developed a predictive model to determine polymer properties and subsequently used it to synthesise polymers with medically desirable properties. We also show that synthetic data generated from small datasets can match the performance of equivalent-sized real data, thereby saving considerable experimental time otherwise spent on data generation in real-world settings by leveraging synthetic data instead. Finally, we pre-trained VECT-GAN on the molecular chemistry of drug-like molecules from the ChEMBL database as a knowledge distillation approach to optimise its utility for pharmaceutical datasets involving drugs. We demonstrate its effectiveness by applying it to two additional pharmaceutical datasets, where it successfully improved regression model performance.

# Results and Discussion

## VECT-GAN - A new synthetic data generator combining variational autoencoders with generative adversarial networks

To address the challenge of data scarcity in pharmaceutical research, we defined a novel model architecture that we hypothesised would show an increased ability to generate realistic data in this space. Data in pharmaceutical science is characterised by high variability, a high number of discrete and categorical features, and a high feature-to-sample ratio. We surmised that combining the strengths of autoencoders (AE), and generative adversarial networks (GANs) would allow a significant advancement in the ability of a generative model to produce samples that reflect real data, with realistic variation that improves the generalisability of fit models (Figure 1).

Figure 1Tabular Variational AE (TVAE) and Conditional Tabular GANs (CTGAN) commonly outperform other models in augmenting tabular datasets[19-21]. Therefore, we introduce the VECT-GAN model to leverage the benefits of both approaches: the adversarial training power of GANs and the latent data representations derived through a variational autoencoder (VAE). Specifically, the VAE component captures data representations allowing the learning of uncertain distributions for effective sampling, while the GAN component ensures the realism of generated data. By sampling from a latent space learned by the VAE rather than random noise, our model maximises learning from data, which is particularly valuable when working with small datasets. Given that pharmaceutical datasets often contain multiple discrete categorical variables, we adopt the CTGAN architecture for the GAN component, with the VAE decoder acting as the generator within the CTGAN framework. CTGANs have been recognised as state-of-the-art for modelling tabular data due to their ability to handle both discrete and continuous data, multimodal continuous distributions, as well as their training-by-sampling approach, which accounts for class imbalances within discrete columns. This is particularly important for small datasets where minority classes are prevalent[19,22].

The quality of all generated datasets was evaluated against the real data (Table S1). Data heterogeneity was prioritised over similarity to the real data, as a broader exploration of the feature space can enhance regression model performance. As a result, synthetic data quality was assessed primarily based on model performance (the ability to predict for unseen real data) rather than strict similarity to the real data.

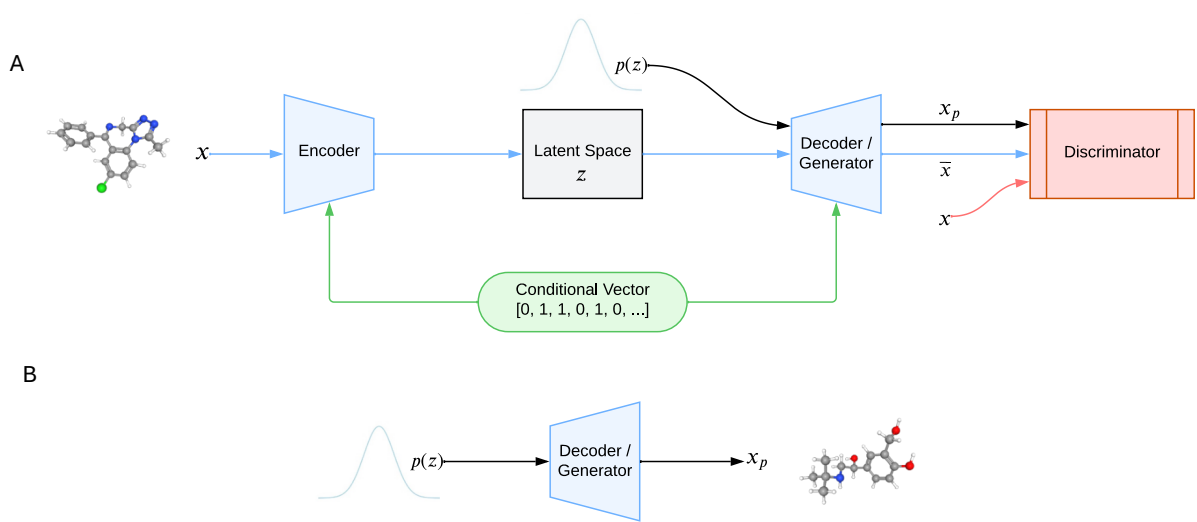

*Figure 1: VECT-GAN architecture. (A) The fused autoencoder and GAN structure: the VAE component is fed the data x and a conditional vector for discrete columns. This input is encoded into the latent space z. The latent representation z is then input into the decoder of the AE, which also serves as the generator of the GAN, to produce reconstructed data, $\overline{x}$. Additionally, Gaussian noise sampled from the latent space distribution, p(z), is input into the decoder to generate synthetic data, $x_p$. The discriminator is trained to differentiate between x, and $\overline{x}$ and $x_p$. (B) During inference, random noise sampled from the latent space distribution p(z) is input into the decoder to generate new synthetic data, $x_p$.*

We next devised a pipeline for generating and assessing the accuracy and use cases of these models on pharmaceutical data (Figure 2). First, real data is split into an 80/20 train and external validation set. The training data is then used to generate synthetic data using our VECT-GAN or one of four other state-of-the-art generative models: a Gaussian copula, a copula GAN (CopulaGAN), a CTGAN or a tabular VAE (TVAE). These models encompass a range of probabilistic and Deep Learning (DL) techniques[23]. A synthetic dataset 3-5 times the size of the training set is generated and concatenated with the original training data to form a new, expanded training set. A model is then fit to the expanded training set, and hyperparameter tuning is conducted using a nested cross-validation approach. This consists of an outer and an inner 5-fold cross-validation loop. In each outer fold, the dataset is split into a training set and a test set. The training set from the outer fold is then used for hyperparameter tuning within the inner fold. Within the inner loop, Bayesian Optimisation is used to select the optimal hyperparameters. Once the best hyperparameters are selected, the model is retrained on the entire training set and tested on the corresponding outer fold using these optimised hyperparameters; this process is repeated across all outer folds.

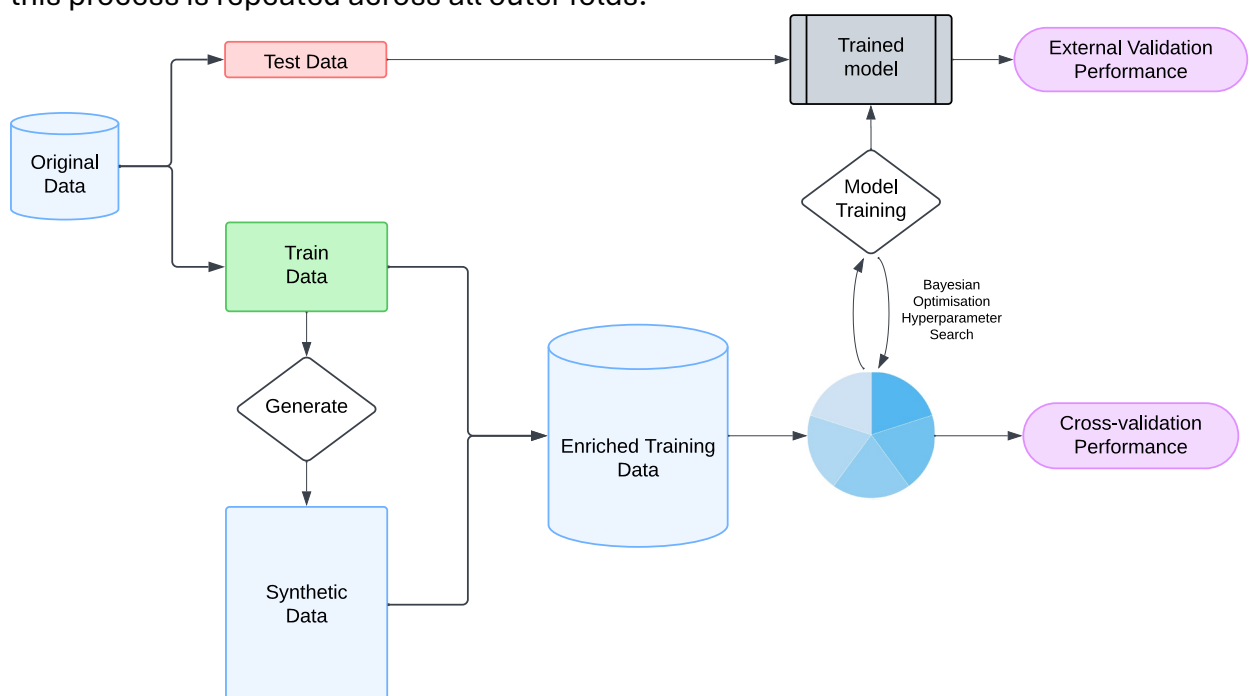

*Figure 2: Pipeline used to generate and evaluate synthetic data.*

## VECT-GAN significantly improves modelling ability in a real world pharmaceutical application

To validate our pipeline in a real experimental system, we chose to study a subset of polymeric materials, mucoadhesive polymers, with strong potential for clinical use in drug delivery but complex discovery processes. Mucoadhesive polymers are materials that attach to mucosal surfaces such as those in the mouth and intestines. By loading these materials with drug molecules, it is possible to significantly increase drug uptake in specific tissues of a patient such as the oesophagus due to increasing the retention time – the amount of time the drug is in contact with epithelial barrier[24]. These materials can be generated using several different base polymers with different manufacturing parameters. Furthermore, chemical modification is often undertaken to enhance adhesion, such as the conjugation of thiol-bearing motifs which bind to proteins present on the mucosa. These materials are highly time-consuming to generate and characterise, with the assessment of adhesion carried out *ex vivo* by attaching the polymer to a piece of mucosal tissue on a spinning paddle to measure adhesion time[25].

We reviewed the literature to collect parameters and adhesion times for published mucoadhesive polymers and thiol-bearing derivatives. This resulted in a dataset of 79 polymer parameter sets and associated adhesion times from 8 publications [26-34]. We used this to train an extreme gradient boosting (XGBoost) regressor to predict adhesion time based on the base polymer, thiolation type, number of thiol groups, the pH at which the polymers were made and their preparation method.

Mucoadhesive polymers are complex to optimise, and they have not been explored well beyond a small set of iterative polymers[35]. Our initial model trained on the collected data alone performed extremely poorly.

We applied our augmentation methodology to the data and evaluated the performance of different models. All models, except for the Gaussian copula, showed improved performance. VECT-GAN emerged as the best model - significantly improving accuracy compared to CopulaGAN, Gaussian copula and training on real data alone (Figure 3A, S3A). This enabled the training of predictive models, halving the MAE scores compared to training on real data. We also assessed combining multiple generated datasets and found that this did not enhance model performance beyond the best individual model, likely due to noise from less effective models impacting overall performance. This improved performance was also observed in the external validation set (Figure 3B, S3B).

To evaluate the robustness of our model, we assessed its ability to differentiate between polymers with high and low adhesion times. Using the model, we selected nine new polymers predicted by our model to exhibit high or low adhesion times and prepared them through variation of process parameters (i.e. lyophilisation/precipitation, pH, polymer grade) and/or synthesis through carbodiimide coupling of cysteine. We then evaluated the model's performance in predicting adhesion times. Our model remained robust when trained on data generated by VECT-GAN, consistently outperforming models trained on other datasets (Figure 3C, S3C). It was found that the model performed better for polymers with low adhesion times compared to those with high adhesion times (Figure 3D). The lower performance for predicting high adhesion times is likely due to the model's inability to predict adhesion times exceeding 10 hours. This limitation is attributed to the fact that many studies from which the training dataset was compiled stopped measuring adhesion time after 10 hours. As a result, the model's predictions are biased towards this value when the actual adhesion time is longer. Subsequent analysis using Shapley Additive Explanations (SHAP) revealed that the number of thiol groups (TG) and pH were the most influential factors in determining adhesion time (Figure 3E). This finding is consistent with existing literature[24,36], indicating that the synthetic data generated using the VECT-GAN preserves the integrity of relationships between features present in the real data.

This use case exemplifies the applicability of our model for pharmaceutical data. It not only aligns with a typical pharmaceutical pipeline but also addresses the challenges inherent in such data, which often includes a mix of discrete and categorical variables, small sample sizes, and significant noise. The model's ability to effectively handle these complexities highlights its broad utility and potential in the pharmaceutical field.

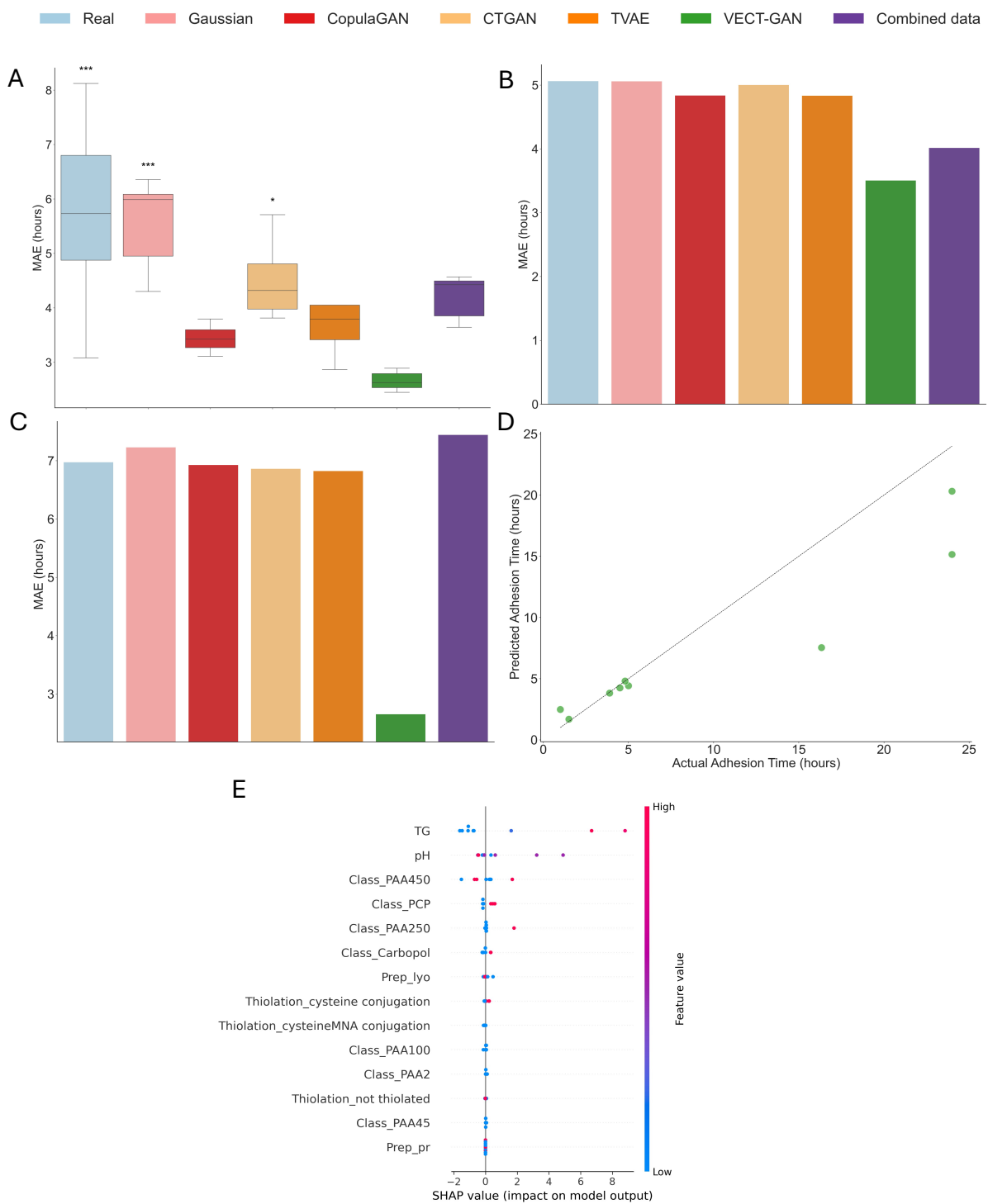

Figure 3: Mucoadhesive polymer results. (A) Nested cross-validation scores measured; statistical significance was determined using a one-way analysis of variance (ANOVA) with post-hoc Tukey's test (* indicates $p < 0.05$, ** indicates $p < 0.01$, and *** indicates $p < 0.001$). Statistical significance is shown only in comparison to VECT-GAN; for the full statistical analysis, please see Table S3. (B) Model performance on the external validation set. (C) Model performance on newly synthesised polymers. (D) Actual versus predicted values for synthesised polymers. (E) SHAP plot for the XGBoost model trained on VECT-GAN data.

# VECT-GANs consistently improve predictive model accuracy in varied pharmaceutical datasets

To further evaluate the impact of data augmentation, and benchmark our VECT-GANs on model performance, we selected additional datasets from the literature to test.

We chose datasets with an increased number of features compared to our initial validated study of mucoadhesive polymers. The first dataset comprised 668 unique self-emulsifying drug delivery systems (SEDDS) of 20 poorly soluble drugs[37]. 27 features comprising physicochemical properties of the drugs and the compositional makeup of the nanoparticles were used to predict particle size. The second dataset comprised 100 drugs and their measured solubilities[38,39]. Solubility is a notoriously difficult challenge for ML. The drugs were represented using their Simplified Molecular Input Line Entry System (SMILES) annotations, from which the drug Morgan Fingerprint (MFP) was generated to predict solubility. We reduced these features down to 10 using principal component analysis to balance the feature-to-sample ratio. Using the MFP is common in pharmaceutical research, enabling us to demonstrate that our methods apply to standard and challenging use cases involving small molecule drugs.

In both cases, augmentation improved predictive power significantly compared to the original dataset, with models trained on VECT-GAN-generated synthetic data having the lowest error (Figure 4A-E, S2). t-distributed Stochastic Neighbour Embedding (t-SNE) applied to the training data, VECT-GAN synthetic data, CTGAN synthetic data (the second-best performing model), and external validation data revealed that synthetic data overlaps with the real data while also exploring a broader state space, providing an explanation for the improved performance observed (Figure 4F). Additionally, the VECT-GAN data is more closely aligned with the distribution of the real and validation data compared to the other datasets. In contrast, the CTGAN data exhibits a broader distribution around the edges of the data space, which likely contributes to VECT-GAN's superior model performance.

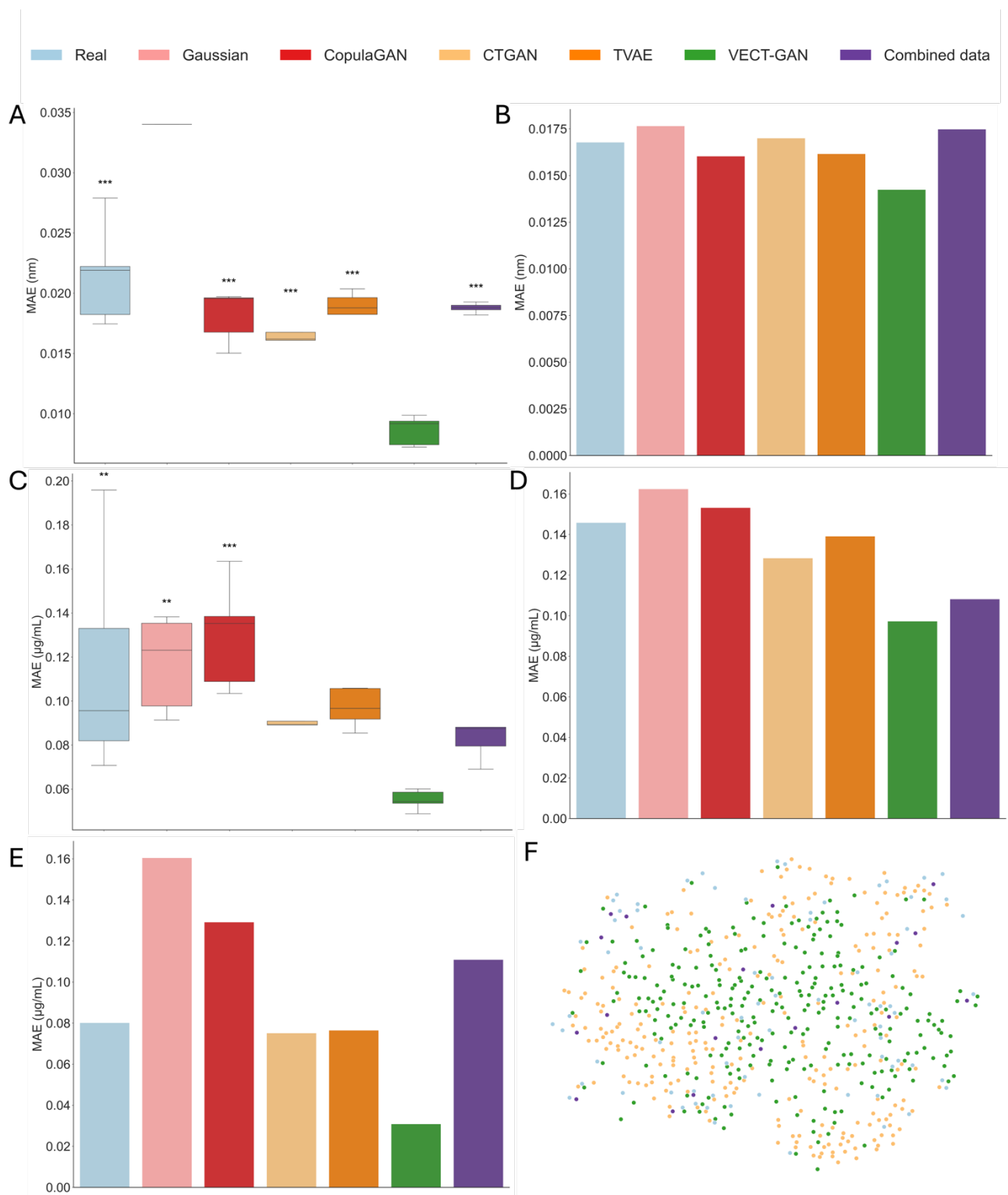

*Figure 4: VECT-GAN performance on SEDDS and solubility datasets. (A) Nested cross-validation scores for the SEDDS dataset; (B) model performance on the external validation set for the SEDDS dataset; (C) nested cross-validation scores for the solubility dataset; (D) model performance on the external validation set for the solubility dataset; (E) model performance on a second external validation set for the solubility dataset, later published by the same authors; (F) t-SNE plots of VECT-GAN, CTGAN (second-best performing model), and real data. Statistical significance was determined using a one-way ANOVA with post-hoc Tukey's test (\* indicates $p < 0.05$, \*\* indicates $p < 0.01$, and \*\*\* indicates $p < 0.001$), with significance shown only in comparison to VECT-GAN; for full statistical details, please see Table S3.*

# Data subsetting demonstrates the utility of augmentation in pharmaceutical science

Having established that this pipeline successfully enhances model performance, we sought to compare the effectiveness of synthetic data with that of real data. To do this, we subset a larger dataset and applied augmentation. The performance of the synthetic data was compared against both the smaller and larger real datasets. We analysed a dataset of 1,982 fast disintegrating tablets[40]. This dataset includes 22 physicochemical properties of drugs and powders and was used to predict disintegration time. Initially, a subset of 80 tablets was taken and augmented to 400 data points, and the model's performance was compared to an equal number of real data points. In the nested cross-validation, all generative models enhanced the performance of the regression model, with the VECT-GAN showing the best overall performance, significantly improving performance compared to the real data (Figure 5A, S5A). However, when comparing real data of equivalent size to synthetic data, the real data demonstrated better model performance, indicated by a lower MAE, although this was not statistically significant. On the external validation set, the VECT-GAN model again outperformed all others, including models trained on the real dataset of equivalent size (Figure 5B, S5B). These analyses demonstrated that the synthetic data exhibited performance similar to that of real data of equivalent size.

We next sought to validate the performance of augmentation on different subset sizes of data. We subset the data into varying sizes, and augmented each in turn, comparing it to real data. External validation shows that model performance for synthetic data was consistently superior to real data points, and VECT-GANs was the best model for all subsets except for one (Figure 5C, S5C). The greatest improvement in model performance was for the smallest dataset where the data was likely too sparse for robust model performance and where data regularisation is most beneficial[41]. Beyond this, improvements were less pronounced. The trend followed by the synthetic data is similar to that observed by the real data, highlighting that this could also be used as a tool to highlight when more real data should be generated. Model performance started to converge for the largest datasets, suggesting that while data augmentation can be beneficial, it is less effective when the dataset is sufficiently large for ML model development, and as regularisation becomes less important.

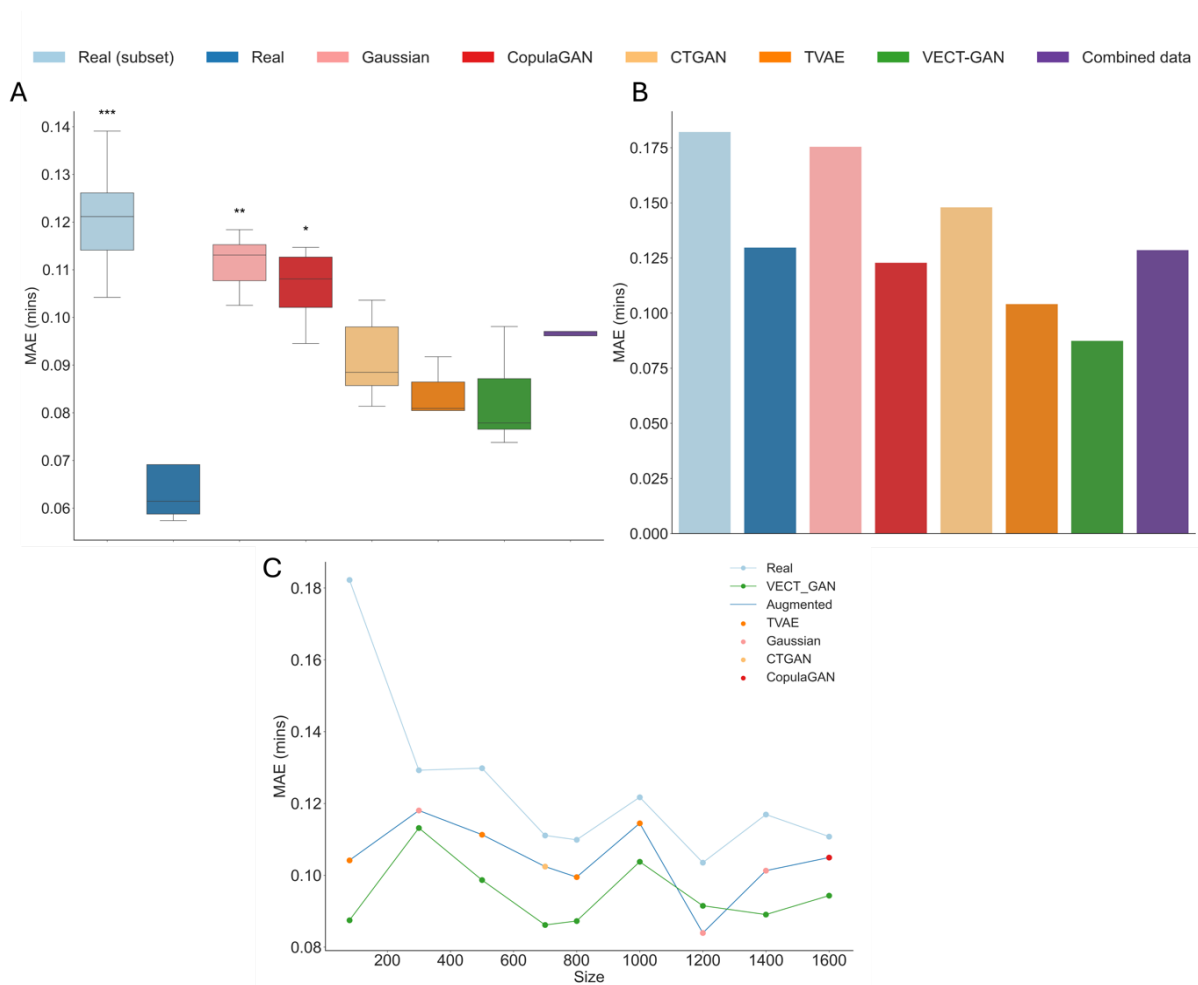

*Figure 5: Comparison of synthetic and real data of the same size for the fast disintegrating tablets dataset.(A) Nested cross-validation scores for 80 real data points, an augmented dataset of 480 data points, as well as the equivalent amount of real data. Statistical significance was determined using a one-way ANOVA with post-hoc Tukey's test (\* indicates $p < 0.05$, \*\* indicates $p < 0.01$, and \*\*\* indicates $p < 0.001$), significance shown only in comparison to VECT-GAN, for the full statistical analysis, see Table S3. (B) Model performance on the external validation set for the same data. (C) Model performance of real and synthetic datasets for different subsets of real data measured against the external validation set (the size refers to the number of real data points in the dataset; for synthetic data, this means that amount of real data plus additional synthetic data); model performance is shown for the real data, VECT-GAN synthetic data, and the other best-performing generative model at each subset size.*

# VECT-GAN trained on ChEMBL outperforms other generative models

Since VECT-GAN outperformed other generative models in producing synthetic data, we sought to further adapt it for use with drugs and pharmaceutical materials. To achieve this, we downloaded a database of approximately 2.5 million drug-like molecules from ChEMBL[42]. The molecular descriptors were extracted using RDKit and used to pre-train VECT-GAN to generate drug descriptors, which are commonly applied in pharmaceutical ML. We evaluated the generated data against a hold-out test set of 100,000 drugs, as shown in Table S2, and demonstrated that it successfully generated data similar to the real data.

We hypothesised that this pre-trained model could support data generation for small datasets. This knowledge distillation provides a more informed starting point for generating synthetic data by leveraging a pharmaceutical/molecular structure foundation. This approach is particularly valuable when working with very small datasets. To test this, we downloaded two datasets from The Therapeutic Data Commons[43] - one predicting the Caco-2 permeability of drugs[44] and another predicting the hydration-free energy of drugs[45]. A subset of 50 drugs was randomly sampled from each dataset. We utilised the pre-trained VECT-GAN model weights, excluding the input and output layers (which were modified to accommodate the additional prediction target dimension), and fine-tuned the model to produce synthetic data. We used the re-trained model to generate 250 new data points, which we then compared to an equivalent amount of real data. For both datasets, VECT-GAN data significantly improved model performance compared to real data and other generative models (Figure 6A, B). However, model performance with synthetic data was slightly lower than with the equivalent amount of real data, although this difference was not statistically significant for Caco-2. Furthermore, models trained on VECT-GAN data produced predictions that aligned more closely with the actual values than those trained on real data or data from other generative models. (Figure 6C, D) These results further demonstrate the utility of VECT-GAN, particularly for small datasets, where the pre-trained model can leverage its understanding of drug descriptors to generate meaningful data, thereby improving regression model performance.

Finally – in order to allow others to effectively use VECT-GAN, we have made it available as a pip package, including a VECT-GAN pretrained on ChEMBL at: https://pypi.org/project/vect-gan/

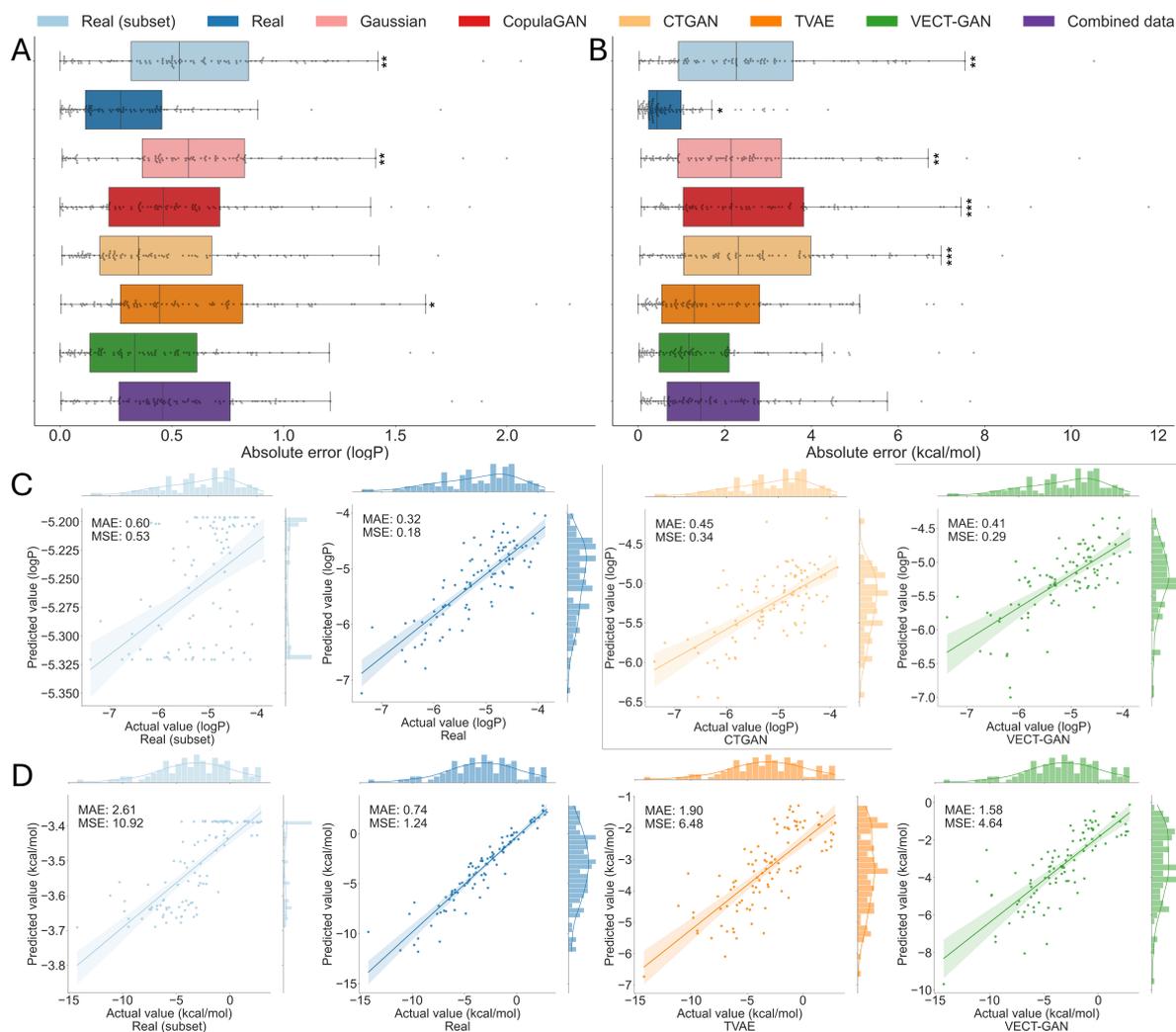

*Figure 6: Model performance of the pre-trained ChEMBL VECT-CTGAN model retrained on two datasets - scores are shown for a subset of real data, augmented data, and real data of the same size as the augmented data. (A) Absolute errors for the model on the test set for the Caco-2 permeability dataset. (B) Absolute errors for the model on the test set for the free solvation dataset. (C) Model predictions on the test set for the real data and top-performing models for the Caco-2 permeability dataset. (D) Model predictions on the test set for the real data and top-performing models for the free solvation dataset, Statistical significance was determined using a one-way ANOVA with post-hoc Tukey's test (\* indicates $p < 0.05$, \*\* indicates $p < 0.01$, and \*\*\* indicates $p < 0.001$), with significance shown only in comparison to VECT-GAN (for the full statistical analysis, refer to Table S4).*

# Conclusion

Herein, we present a novel generative model, VECT-GAN, designed to address the challenge of small datasets in pharmaceutical data science. Through its application to a real-world system which we experimentally validated, datasets, we demonstrate its capability to generate data from diverse sources and its effectiveness in handling small, noisy, multimodal datasets with discrete and continuous columns - features characteristic of pharmaceutical datasets. We further apply VECT-GAN regularisation to a number of existing pharmaceutical datasets to demonstrate its utility. To further adapt VECT-GAN for pharmaceutical use, we pre-trained it on the entire ChEMBL drug-like molecule dataset, enabling knowledge distillation for pharmaceutical tasks and enhancing its utility for small datasets. Overall we developed a novel pipeline for integrating synthetic data into regression models, demonstrating a significant improvement in model performance in almost all of our applied use cases. This approach outperforms state-of-the-art generative models across a variety of pharmaceutical tasks. This work represents the first step towards establishing generative models as a standard tool for regularising data in tabular datasets within pharmaceutical research.

# Methods

## Materials

Poly(acrylic acid) (PAA) with molecular weight 450 and 250 kg/mol (PAA450 and PAA250, respectively) was purchased from Sigma-Aldrich (UK). Noveon® Polycarbophil and Carbopol 974P were supplied from Lubrizol (U.K.) without charge. 1-ethyl-3-(3-dimethylaminopropyl) (EDAC) and L-cysteine were purchased from Sigma-Aldrich (U.K.).

## Data

Details of the different datasets used can be found in Table 1.

| Dataset | Size | Input | Features | Target | Source |
|---|---|---|---|---|---|
| **Mucoadhesive polymers** | 79 | Polymer properties | 5 | Retention time | [26-34] |
| **SEDDS** | 668 | Properties of the drug and the nanoparticles (compositional data) | 26 | Nanoparticle size | [37] |
| **Solubility** | 100 | Drug physiochemical properties | 210 | Drug solubility | [38] |
| **Fast disintegrating tablets** | 1982 | Drug physicochemical properties and tablet/powder properties | 22 | Disintegration time | [40] |
| **Caco-2 permeability** | 818 | Drug physiochemical properties | 210 | Caco-2 permeability | [44] |
| **Free Solvation Database** | 578 | Drug physiochemical properties | 210 | Hydration free energy | [45] |

*Table 1: summary of the datasets used.*

## Generative models

Four generative models were used in this study, TVAE, CopulaGAN, CTGAN and Gaussian Copula. These models were run using Synthetic Data Vault (Version 1.4.0)[46] on Python (Version 3.10.4) on a Windows desktop (Operating System: Windows 11; Processor: AMD Ryzen Threadripper 7960X 24-core 4.2 GHz; RAM Memory: 128 GB, GPU: RTX 4090 24 GB). The complete code is available in the GitHub repository (https://github.com/y-babdalla/vect_gan). The hyperparameters used for data generation can be found in Table S5.

## VECT-GANs

A hybrid VECT-GAN and CTGAN model was developed. The VAE component provides a structured latent space, enabling the learning of uncertain distributions for effective sampling. The GAN improves the quality of generated data through adversarial training, while the conditional tabular component facilitates conditional generation and sampling-based training, effectively handling discrete columns. The encoder, decoder and discriminator networks consisted of multiple Residual layers. This layer applies a linear transformation, batch normalisation, and a *ReLU* activation, with a skip connection that concatenates the input with the transformed output:

$$Residual(x) = ReLU(BatchNorm(Linear(x))) \,||\, x$$

The VAE Encoder network is multiple residual blocks, followed by a final linear layer that reduces the dimensionality to the latent space.

Let $x \in \mathbb{R}^{d_{input}}$ represent the input data, $c$ the conditional vector, $h$ the hidden representation after residual blocks, and $z \in \mathbb{R}^{d_{latent}}$ the latent variable output, after the reparameterisation trick:

$$h = Residual_n(\ldots Residual_1(x + c))$$
$$\mu = Linear(h)$$
$$\log \sigma^2 = Linear(h)$$
$$z = \mu + \varepsilon \cdot e^{0.5 \cdot \log \sigma^2}, \varepsilon \sim N(0, I)$$

Where $\mu$ is the mean vector of the latent distribution, $\log \sigma^2$ represents the log variance of the latent distribution and $\varepsilon$ is a sample from a standard normal distribution used to apply the reparameterisation trick.

The VAE Decoder/GAN generator network is structured symmetrically to the encoder. It expands the latent variable z back to the original data dimension:

$$\hat{x} = Residual_1(\ldots Residual_n(z + c))$$

The Discriminator network receives real, reconstructed or generated data, and outputs a score indicating the likelihood that the input is real. The discriminator's structure can be represented as follows:

$$D(x) = Linear_m(\ldots LeakyReLU(Dropout(Linear_1(x)))\ldots)$$

To enforce conditional generation, a conditional vector was incorporated into the VAE and GAN training processes, with conditional cross-entropy loss calculated. The VAE loss combines reconstruction loss, conditional cross-etropy loss, and Kullback-Leibler (KL) Divergence. The generator (or decoder) loss is a combination of the VAE loss and the discriminator score for both fake and reconstructed data. The discriminator loss, includes scores for real, reconstructed, and fake data, with an added gradient penalty to enforce Lipschitz continuity, to help prevent mode collapse[47]. Further details of model training can be found in Algorithm 1. The hyperparameters for VECT-GAN can be found in Table S5.

**Algorithm 1: VECT-GAN Training**

**Initialise parameters** $\theta_{Enc}, \theta_{Dec}, \theta_{Dis}$

**Repeat**
$Z \leftarrow Enc(X + c)$
$\mathcal{L}_{prior} \leftarrow D_{KL}(q(Z|X) \parallel p(Z))$
$\hat{X} \leftarrow Dec(Z + c)$

$$\mathcal{L}_{Reconstruction} = -\sum_{j=1}^{d_{discrete}} X_j log(\hat{X}_j) + \sum_{j=1}^{d_{continuous}} (X_j - \hat{X}_j)^2$$

$\mathcal{L}_{Conditional} \leftarrow Cross\ Entropy\ Loss\ for\ Discrete\ Columns\ in\ Conditioned\ Data$
$\mathcal{L}_{VAE} = \mathcal{L}_{Reconstruction} + \mathcal{L}_{prior} + \mathcal{L}_{conditional}$
$Z_p \sim samples\ from\ prior\ \mathcal{N}(0, I)$
$X_p \leftarrow Dec(Z_p + c)$
$\mathcal{L}_{Generator} = \mathcal{L}_{VAE} + 0.5(-\mathbb{E}[D(X_p)] - \mathbb{E}[D(\hat{X})])$

**Repeat n times for each loop:**
$$\mathcal{L}_{Discriminator} = -\mathbb{E}[D(X)] + 0.5(-\mathbb{E}[D(X_p)] - \mathbb{E}[D(\hat{X})]) + \lambda\mathbb{E}\left[\left(\left\|\nabla_{X_p}D(X_p)\right\|_2 - 1\right)^2\right] +$$
$$\lambda\mathbb{E}\left[\left(\left\|\nabla_{\hat{X}}D(\hat{X})\right\|_2 - 1\right)^2\right]$$

**Update parameters**
$$\theta_{Enc} \leftarrow \theta_{Enc} - \nabla_{\theta_{Enc}}\mathcal{L}_{VAE}$$
$$\theta_{Dec} \leftarrow \theta_{Dec} - \nabla_{\theta_{Dec}}\mathcal{L}_{Generator}$$
$$\theta_{Dis} \leftarrow \theta_{Dis} - \nabla_{\theta_{Dis}}\mathcal{L}_{Discriminator}$$

**Until convergence or max epochs**

Where $\theta$ is the network parameters, $X$ is the input data, $c$ is the conditional vector, $Z$ is the latent space, $\hat{X}$ reconstructed input, $\mathcal{L}$ is the loss, $D_{KL}(q(Z|X) \parallel p(Z))$ is the KL-divergence between the approximate posterior and prior distribution, $X_p$ is the vector sampled from the prior distribution, $X_p$ is the generated sample, **Enc(·)** is the Encoder network function, **Dec(·)** is the Decoder (Generator) network function and **D(·)** is the Discriminator network function.

## Regression Model training and tuning

Model training and hyperparameter tuning were performed using a 5-fold nested cross-validation approach. In the outer loop, the data was split into five segments. For each fold, one segment was used for testing, while the remaining data was used for training. Within the inner loop, the training data underwent another 5-fold cross-validation for hyperparameter tuning guided by Bayesian optimisation through Optuna, optimising the mean squared error. Once the optimal hyperparameters were identified, they were applied to the outer loop to test on the external validation set. Model performance was measured using MAE and MSE. The model used for this analysis was XGBoost (Version 1.6.2), with evaluation and hyperparameter tuning carried out using the Scikit-learn Python package (Version 1.1.3). The hyperparameters search space can be found in Table S6.

### Morgan Fingerprint and molecular descriptors generation

The MFP (2048 bits, radius 2) and molecule descriptors were computed from the SMILES annotation using RDkit (Version 2022.9.5).

### Dimensionality reduction

Dimensionality reduction using PCA and t-SNE was carried out using the Scikit-learn (Version 1.1.3).

### Explainability analysis

SHAP was used to identify the impact of different features on model predictions. This was conducted using the Python SHAP package (Version 0.42.1).

## Statistics

All statistical analyses were performed using the SciPy package (Version 1.11.1) in Python. A one-way analysis of variance (ANOVA) with post-hoc Tukey's tests was used to assess significance, with a threshold of $p < 0.05$.

## Mucoadhesive tests

### Polymer synthesis

PAA was hydrated in demineralized water (1 g per 80 mL) and the pH of the solution was adjusted to 6 by the addition of 5 M NaOH. EDAC was then added with continual stirring. After 20 min incubation at room temperature, L-cysteine was added and the pH was readjusted to 6. Reaction mixtures were incubated for 3 h at room temperature under stirring. The resulting conjugates were isolated by dialysis according to the method described previously for PCP–Cys[48]. Control polymers were prepared and isolated in the same way, however EDAC was omitted during the coupling reaction. After dialysis, the pH of all samples was readjusted to 6 and frozen aqueous polymer solutions were dried by lyophilization. All conjugates and controls were stored at 4 °C until further use. The synthesised polymers are summarised in Table 2.

The degree of thiolation of the polymers was determined by $^1$H NMR ($D_2O$ with NaOH), calculated from the ratio of integrals at 3.9 ppm, relating to the SC$H_2$ of the cysteine moiety, and 1.2 ppm, arising from the C$H_2$ of the acrylic acid backbone (Supplementary Figure.

| Free thiol group content (micromol/g) | Preparation | pH | Class | Thiolation |
|---|---|---|---|---|
| 495 | lyophilized | 5 | PAA450 | cysteine conjugation |
| 340 | lyophilized | 5 | PAA250 | cysteine conjugation |
| 163 | lyophilized | 5 | PCP | cysteine conjugation |
| 0 | precipitated | 5 | Carbopol | not thiolated |
| 0 | lyophilized | 5 | Carbopol | not thiolated |
| 0 | precipitated | 4 | PCP | not thiolated |
| 0 | lyophilized | 4 | PCP | not thiolated |
| 0 | precipitated | 6 | PAA450 | not thiolated |
| 0 | lyophilized | 6 | PAA450 | not thiolated |

*Table 2: properties of polymers synthesised.*

## In vitro mucoadhesion studies

Mucoadhesion studies were performed using a rotating paddle method in a dissolution apparatus. Polymer samples were pressed into 10 mm diameter discs under a compressive force of 5 bar. The polymer discs were pressed onto excised porcine intestinal mucosa that were spanned onto stainless-steel rotating paddles and then completely immersed into 900 mL of 100 mM phosphate buffer solution of pH 6.8. The test was performed at 37°C with a rotational speed of 150 rpm. Adhesion of the polymer discs to the porcine mucosa was observed every 60 mins until the discs were either disintegrated or detached from the mucosa. Each mucoadhesion test was done in triplicate for each polymer disc.

# Supplementary Information

|  | Gaussian | CopulaGAN | CTGAN | TVAE | VECT-GAN |
|---|---|---|---|---|---|
| **Polymer** | | | | | |
| **KS Test** | 0.77 | 0.86 | 0.82 | 0.85 | 0.86 |
| **TV Test** | 0.88 | 0.93 | 0.94 | 0.92 | 0.93 |
| **KL Divergence** | 0.37 | 0.53 | 0.55 | 0.59 | 0.53 |
| **SEDDS** | | | | | |
| **KS Test** | 0.80 | 0.84 | 0.85 | 0.84 | 0.84 |
| **KL Divergence** | 0.26 | 0.60 | 0.61 | 0.62 | 0.60 |
| **Solubility** | | | | | |
| **KS Test** | 0.90 | 0.90 | 0.88 | 0.92 | 0.90 |
| **KL Divergence** | 0.26 | 0.31 | 0.30 | 0.37 | 0.31 |
| **Fast disintegrating tablets** | | | | | |
| **KS Test** | 0.83 | 0.85 | 0.85 | 0.86 | 0.85 |
| **KL Divergence** | 0.23 | 0.43 | 0.44 | 0.53 | 0.43 |
| **Caco-2 permeability** | | | | | |
| **KS Test** | 0.89 | 0.77 | 0.80 | 0.84 | 0.77 |
| **KL Divergence** | 0.50 | 0.48 | 0.51 | 0.67 | 0.48 |
| **Free solvation database** | | | | | |
| **KS Test** | 0.87 | 0.81 | 0.81 | 0.83 | 0.81 |
| **KL Divergence** | 0.53 | 0.60 | 0.59 | 0.75 | 0.60 |

**Table S1:** Evaluation of synthetic data quality using three metrics: the Kolmogorov-Smirnov statistic (KS Test) for continuous columns, the Total Variation Distance test (TV Test) for discrete columns - both ranging from 0 to 1, with 1 indicating perfect similarity - and the Kullback-Leibler (KL) divergence, which measures the divergence between two probability distributions; a lower score indicates greater similarity.

| CHEMBL Performance | |
|---|---|
| **KS Test** | 0.93 |
| **KL Divergence** | 0.75 |

**Table S2:** Evaluation of the synthetic data for the model trained on the entire of CHEMBL, tested on a hold-out test set of 100,000 drugs.

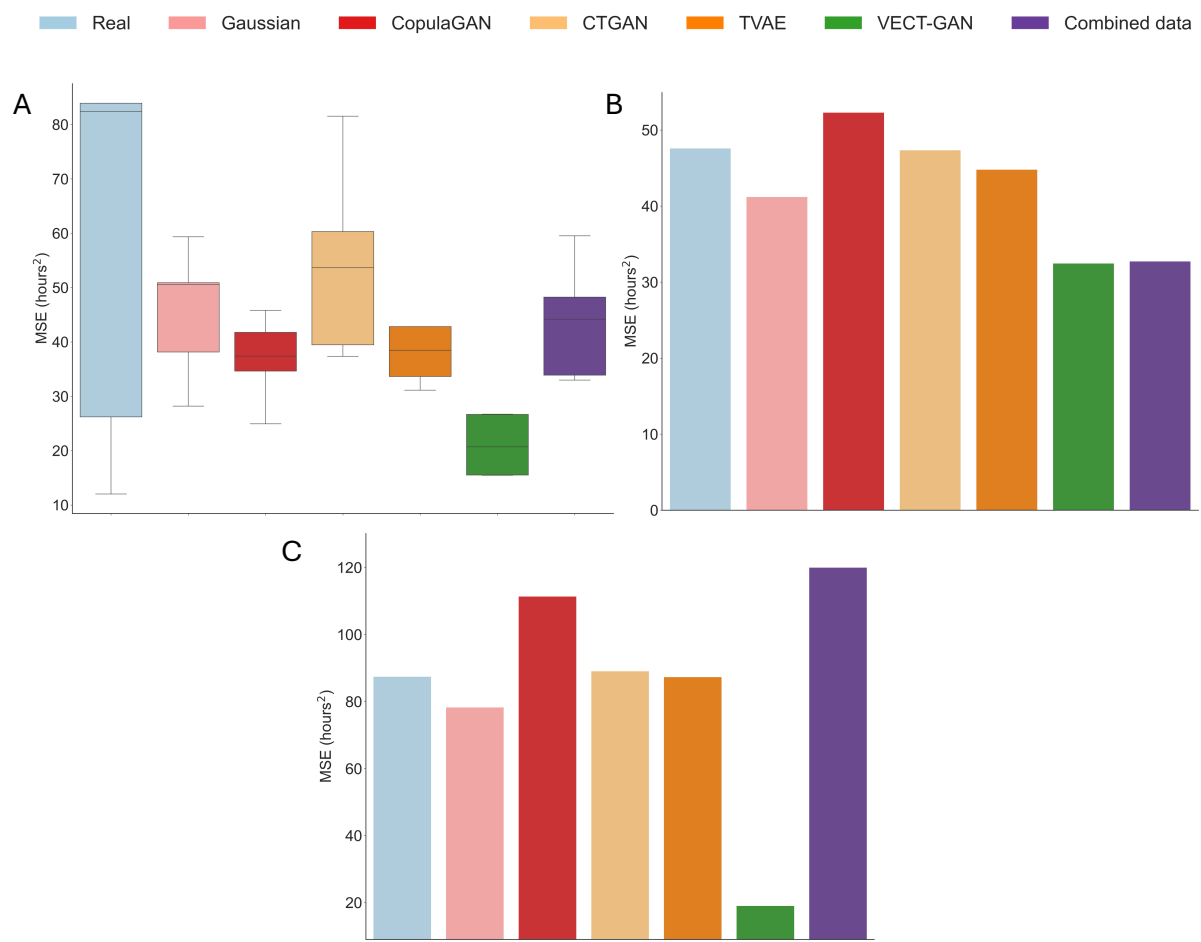

Figure S1: Additional mucoadhesive polymer metrics. (A) Nested cross-validation scores. (B) Model performance on the external validation set. (C) Model performance on newly synthesised polymers.

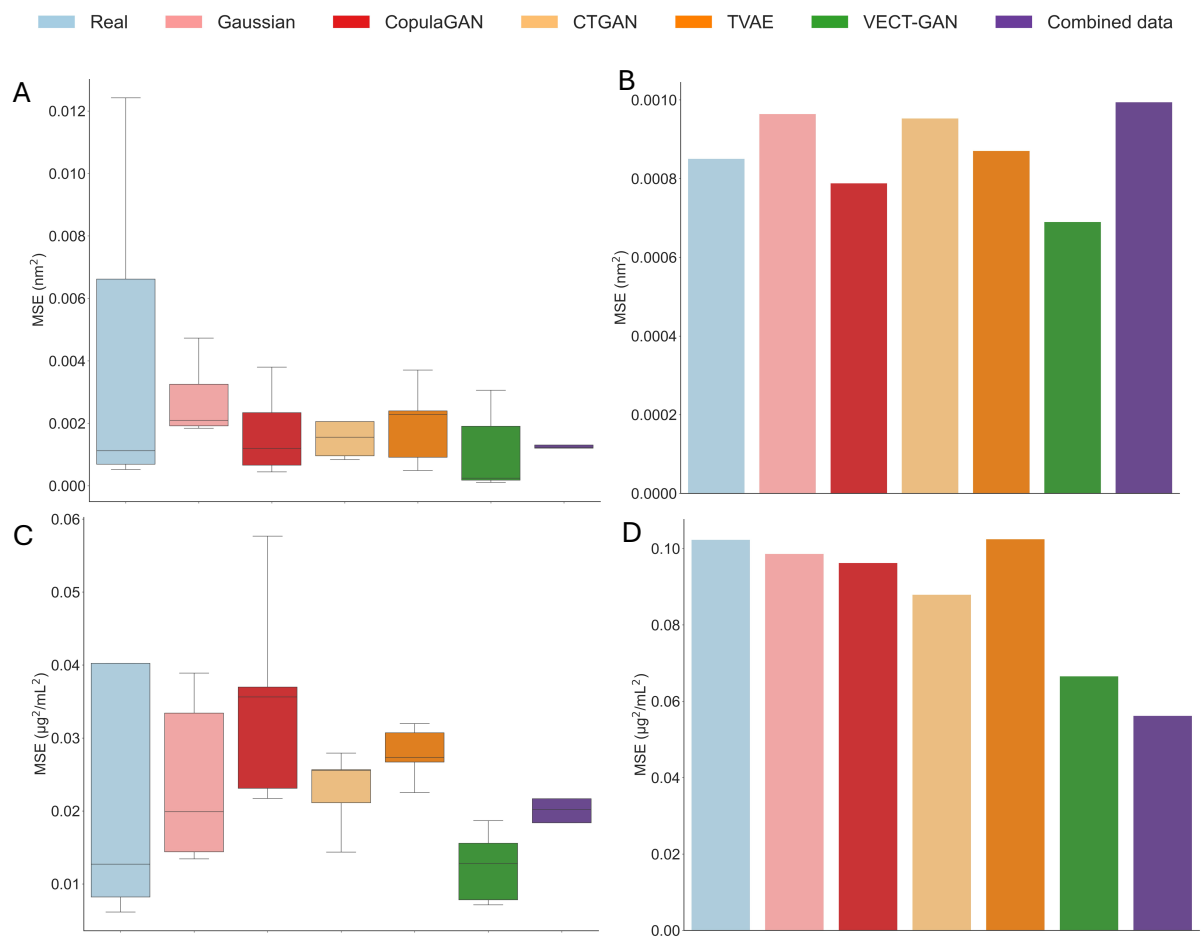

Figure S2: Additional VECT-GAN performance metrics on SEDDS and solubility datasets. (A) Nested cross-validation scores for the SEDDS dataset; (B) model performance on the external validation set for the SEDDS dataset; (C) nested cross-validation scores for the solubility dataset; (D) model performance on the external validation set for the solubility dataset.

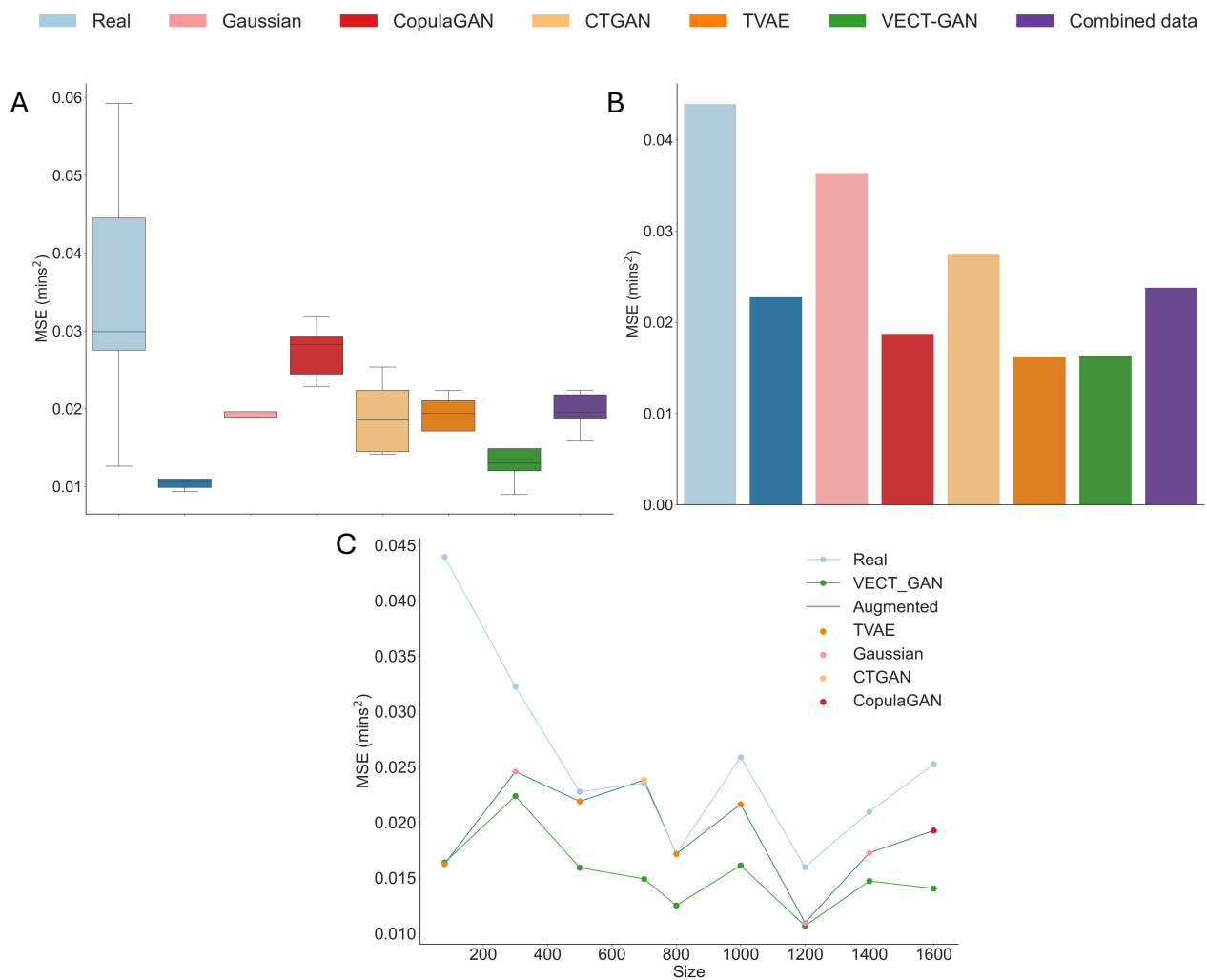

Figure S3: Additional metrics for the comparison of synthetic and real data of the same size for the fast disintegrating tablets dataset. (A) Nested cross-validation scores for 80 real data points, an augmented dataset of 480 data points, as well as the equivalent amount of real data. (B) Model performance on the external validation set for the same data. (C) Model performance of real and synthetic datasets for different subsets of real data measured against the external validation set (the size refers to the number of real data points in the dataset; for synthetic data, this means that amount of real data plus additional synthetic data); model performance is shown for the real data, VECT-GAN synthetic data, and the other best-performing generative model at each subset size.

| Dataset | Metric | Model 1 | Model 2 | p value |
|---|---|---|---|---|
| **Mucoadhesive polymers** | MAE | CopulaGAN | Gaussian | 0.0189 |
| | MAE | CopulaGAN | Real | 0.0086 |
| | MAE | CTGAN | VECT GAN | 0.0475 |
| | MAE | Gaussian | VECT GAN | 0.0006 |
| | MAE | Real | TVAE | 0.046 |
| | MAE | Real | VECT GAN | 0.0003 |
| **SEDDS** | MAE | Combined Data | Gaussian | <0.00001 |
| | MAE | Combined Data | VECT GAN | <0.00001 |
| | MAE | CopulaGAN | Gaussian | <0.00001 |
| | MAE | CopulaGAN | VECT GAN | <0.00001 |
| | MAE | CTGAN | Gaussian | <0.00001 |
| | MAE | CTGAN | Real | 0.0103 |
| | MAE | CTGAN | VECT GAN | <0.00001 |
| | MAE | Gaussian | Real | <0.00001 |
| | MAE | Gaussian | TVAE | <0.00001 |
| | MAE | Gaussian | VECT GAN | <0.00001 |
| | MAE | Real | VECT GAN | <0.00001 |
| | MAE | TVAE | VECT GAN | <0.00001 |
| **Solubility** | MAE | CopulaGAN | VECT GAN | 0.0005 |
| | MAE | Gaussian | VECT GAN | 0.0048 |
| | MAE | Real | VECT GAN | 0.0065 |
| **Fast disintegrating tablets** | MAE | Combined Data | Real (subset) | 0.003 |
| | MAE | Combined Data | Real | 0.0017 |
| | MAE | CopulaGAN | Real | <0.00001 |
| | MAE | CopulaGAN | TVAE | 0.0045 |
| | MAE | CopulaGAN | VECT GAN | 0.0108 |
| | MAE | CTGAN | Gaussian | 0.0493 |
| | MAE | CTGAN | Real (subset) | 0.0008 |
| | MAE | CTGAN | Real | 0.006 |
| | MAE | Gaussian | Real | <0.00001 |
| | MAE | Gaussian | TVAE | 0.0005 |
| | MAE | Gaussian | VECT GAN | 0.0012 |
| | MAE | Real (subset) | Real | <0.00001 |
| | MAE | Real (subset) | TVAE | <0.00001 |
| | MAE | Real (subset) | VECT GAN | <0.00001 |
| | MSE | Real (subset) | Real | 0.0032 |
| | MSE | Real (subset) | TVAE | 0.0343 |
| | MSE | Real (subset) | VECT GAN | 0.0128 |

Table S3: p-values for statistically significant differences in performance during nested cross-validation, assessed using ANOVA with post-hoc Tukey's test.

| Dataset | Model 1 | Model 2 | p value |
|---|---|---|---|
| **Caco-2 Permeability** | Combined Data | Real | 0.0059 |
| | CopulaGAN | Real | 0.0086 |
| | Gaussian | Real | <0.00001 |
| | Gaussian | VECT GAN | 0.0047 |
| | Real (Subset) | Real | <0.00001 |
| | Real (Subset) | VECT GAN | 0.0071 |
| | Real | TVAE | <0.00001 |
| | TVAE | VECT GAN | 0.0202 |
| **Free Solvation Database** | Combined Data | CopulaGAN | 0.0425 |
| | Combined Data | CTGAN | 0.0232 |
| | Combined Data | Real | 0.0001 |
| | CopulaGAN | Real | <0.00001 |
| | CopulaGAN | TVAE | 0.0321 |
| | CopulaGAN | VECT GAN | 0.0002 |
| | CTGAN | Real | <0.00001 |
| | CTGAN | TVAE | 0.0171 |
| | CTGAN | VECT GAN | 0.0001 |
| | Gaussian | Real | <0.00001 |
| | Gaussian | VECT GAN | 0.0061 |
| | Real (Subset) | Real | <0.00001 |
| | Real (Subset) | VECT GAN | 0.001 |
| | Real | TVAE | 0.0001 |
| | Real | VECT GAN | 0.0194 |

Table S4: p-values for statistically significant differences in absolute errors for the pre-treined GAN, assessed using ANOVA with post-hoc Tukey's test.

| Model | Parameter | Value |
| --- | --- | --- |
| **CTGAN** | embedding_dim | 128 |
| | generator_dim | (256, 256) |
| | discriminator_dim | (256, 256) |
| | generator_lr | 2e-4 |
| | generator_decay | 1e-6 |
| | discriminator_lr | 2e-4 |
| | discriminator_decay | 1e-6 |
| | batch_size | 500 |
| | discriminator_steps | 1 |
| | log_frequency | True |
| | verbose | False |
| | pac | 10 |
| **CopulaGAN** | embedding_dim | 128 |
| | generator_dim | (256, 256) |
| | discriminator_dim | (256, 256) |
| | generator_lr | 2e-4 |
| | generator_decay | 1e-6 |
| | discriminator_lr | 2e-4 |
| | discriminator_decay | 1e-6 |
| | batch_size | 500 |
| | discriminator_steps | 1 |
| | log_frequency | True |
| | verbose | False |
| | epochs | 300 |
| | pac | 10 |
| **TVAE** | embedding_dim | 128 |
| | compress_dims | (128, 128) |
| | decompress_dims | (128, 128) |
| | l2scale | 1e-5 |
| | batch_size | 500 |
| | epochs | 300 |
| | loss_factor | 2 |
| **VECT-GAN** | embedding_dim | 32 |
| | encoder_dim | (32, 64) |
| | discriminator_dim | (32, 32) |
| | generator_lr | 1e-4 |
| | discriminator_lr | 1e-4 |
| | encoder_lr | 1e-5 |
| | kl_weight | 1 |
| | vae_weight | 1 |
| | discriminator_decay | 1e-5 |
| | generator_decay | 1e-4 |
| | encoder_decay | 1e-5 |
| | epochs | 100 |

|  |  |
|---|---|
| pac | 10 |
| cond_loss_weight | 1 |
| lambda_ | 5 |

**Table S5:** Generative model hyperparameters

| Parameter | Search Space |
|---|---|
| **n_estimators** | 100, 1000 |
| **max_depth** | 1, 20 |
| **learning_rate** | 0.0001, 0.1 |
| **min_child_weight** | 1, 30 |
| **subsample** | 0.1, 1 |
| **colsample_bytree** | 0, 1 |
| **max_delta_step** | 1, 10 |

**Table S6: XGBoost** hyperparameter search space.